\documentclass{article}
\usepackage[preprint]{spconf}
\toappear{Preprint. Under Review}

\usepackage{spconf,amsmath}
\usepackage{graphicx}\graphicspath{ {figures/} }
\usepackage{algorithm, algorithmic}
\usepackage{multirow, multicol, booktabs, microtype}
\usepackage{balance}
\usepackage{hyperref}
\usepackage{bbold}
\usepackage{subcaption}
\usepackage{enumitem}
\setlist{nosep, leftmargin=14pt}
\usepackage{booktabs}
\usepackage{tabularx}
\usepackage{mwe}

\title{Annotation-Efficient Task Guidance for Medical Segment Anything}
\name{Tyler Ward, Abdullah-Al-Zubaer Imran}
\address{Department of Computer Science\\
University of Kentucky, Lexington, KY 40506, USA}

\begin{document}

\maketitle

\begin{abstract}

Medical image segmentation is a key task in the imaging workflow, influencing many image-based decisions. Traditional, fully-supervised segmentation models rely on large amounts of labeled training data, typically obtained through manual annotation, which can be an expensive, time-consuming, and error-prone process. This signals a need for accurate, automatic, and annotation-efficient methods of training these models. We propose \textit{SAM-Mix}, a \textit{novel} multitask learning framework for medical image segmentation that uses class activation maps produced by an auxiliary classifier to guide the predictions of the semi-supervised segmentation branch, which is based on the SAM framework. Experimental evaluations on the public LiTS dataset confirm the effectiveness of SAM-Mix for simultaneous classification and segmentation of the liver from abdominal computed tomography (CT) scans. When trained for 90\% fewer epochs on only 50 labeled 2D slices, representing just 0.04\% of the available labeled training data, SAM-Mix achieves a Dice improvement of 5.1\% over the best baseline model. The generalization results for SAM-Mix are even more impressive, with the same model configuration yielding a 25.4\% Dice improvement on a cross-domain segmentation task. Our code is available at \url{https://github.com/tbwa233/SAM-Mix}.

\end{abstract}


\begin{figure*}[t]
    \centering
    \includegraphics[width=0.9\linewidth]{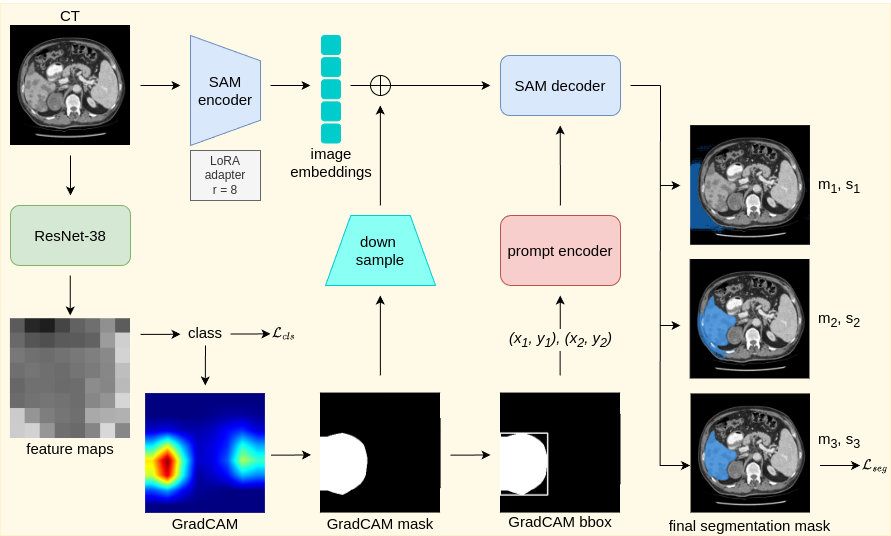}
    \caption{Illustration of the proposed SAM-Mix multitask learning framework. SAM-Mix combines a ResNet-38 classifier for GradCAM generation, automated prompt generation through mask and bounding box extraction, and a SAM-based segmentation pipeline. Low-rank adaptation (LoRA) is used with a rank ($r$) of 8 to make SAM more parameter-efficient. SAM-Mix uses classification-guided attention to produce accurate segmentation masks ($m_1$-$m_3$) with corresponding confidence scores ($s_1$-$s_3$).}
    \label{fig:finaldiagrampng}
\end{figure*}

\section{Introduction}
\label{sec:intro}
Segmentation of targets such as organs, lesions, and tumors is one of the most important tasks in the downstream medical imaging pipeline, playing a vital role in enhancing the diagnostic and monitoring capabilities of clinicians. Traditional U-Net \cite{ronneberger2015u}-based architectures have long been one of the most popular and reliable methods for performing medical image segmentation, and have demonstrated strong performance across a variety of imaging modalities. However, these methods are not without their faults and can struggle with issues like the vanishing gradient problem, lack of understanding of global context in large medical images, and high computational complexity and required computing power.

Recently, the development of foundation models for segmentation such as the Segment Anything Model (SAM) \cite{kirillov2023segment} has opened up new pathways for addressing some of the challenges associated with traditional fully-supervised models. Due to the vast amount of data it was trained on, SAM and its variants have demonstrated strong generalization ability in both zero-shot and few-shot settings. One of the defining characteristics of SAM is that it is a promptable segmentation model, supporting a variety of different prompt formats, including points \cite{xu2023sppnet}, bounding boxes \cite{rahman2024pp}, text \cite{zhang2024evf}, and masks \cite{xie2024masksam}. There has been much focus on the best way to efficiently and accurately prompt SAM-based methods. Fully automated prompt generation is underexplored and has proven largely ineffective when attempted, due to the high task variability across segmentation tasks, particularly in domain-specific fields such as medical imaging \cite{ma2024segment}. As such, the literature surrounding SAM largely focuses on manual or semi-automated prompting, although such prompt generation methods carry several of the same flaws as the manual annotation of training data, in that they can be time-consuming to generate. Additionally, it is not always desirable to increase the parameters in a foundation model such as SAM, as this can lead to issues such as overfitting. 

A potential solution for the issues present in manual or semi-automated prompting of SAM lies in semi-supervised learning (SSL). In SSL, a large number of unlabeled samples are paired with a limited number of labeled samples to learn more efficiently than in a fully-supervised setting. Furthermore, it has been shown that incorporating SSL in a multitask learning framework, where multiple tasks such as segmentation and classification are trained together so that information from one task can be used to improve the other~\cite{imran2020self}. Such multitask learning can improve model generalizability. As an example, in one semi-supervised multi-task model, MultiMix \cite{haque2021multimix}, a learnable connection was created between an auxiliary classification task and a segmentation branch, where saliency information obtained from the classification branch was used to improve the segmentation predictions. In this case, the use case was a binary segmentation problem to determine the presence of pneumonia from chest X-ray images. 

Similar approaches are frequently employed in weakly supervised semantic segmentation (WSSS) models, aiming to produce segmentations based on image-level class labels. Class activation maps (CAMs) are popular in WSSS frameworks and have been incorporated with SAM both for post-processing and zero-shot inference \cite{chen2023segment, chen2023weakly}. However, these methods are vulnerable to noise in the CAMs. Recently proposed S2C \cite{kweon2024sam} attempts to use SAM to address these vulnerabilities within CAM-based WSSS models. S2C attempts to leverage information from SAM to improve the quality of CAMs. While deemed effective, the inverse of this process, using CAMs to enhance the output of SAM, is underexplored. Additionally, S2C's training time is a major downside, as it relies on the computationally expensive process of running SAM's automatic mask generator across the input dataset before the training.

Inspired by S2C~\cite{{kweon2024sam}} andMultiMix~\cite{haque2021generalized}, we propose \textit{SAM-Mix}, a novel multitask learning framework that leverages an innovative task guidance from image-level labels to enhance SAM-based medical image segmentation. Through the connection between these two branches, SAM-Mix offers improved explainability and segmentation accuracy for the specific needs of medical imaging. Our specific contributions are summarized as:
\begin{itemize}
  \item A novel connection between CAMs and SAM, enhancing the segmentation mask predictions through auxiliary classification-guided attention.
  \item Our SAM-Mix model demonstrates impressive performance even when trained on as few as 5 segmentation-labeled abdominal computed tomography (CT) slices. Additionally, we demonstrate SAM-Mix's strong generalizability in a cross-domain evaluation.
  \item Thorough experimentation demonstrating our model's superiority over fully-supervised and semi-supervised models in segmenting liver from the public LiTS and TotalSegmentator datasets.
\end{itemize} \

\section{Methods: SAM-Mix}
\label{sec:methods}
To formulate the problem, we assume a data distribution $p(\mathit{X}, \mathit{Y})$ over $\mathit{D}$ where $\mathit{X} = \{x_1, x_2, \dots, x_N\}$ is a set of abdominal CT slices and $\mathit{Y} = \{y_1, y_2, \dots, y_N\}$ is the set of corresponding ground truth segmentation maps. A network $\mathit{G}_{\phi}$ is created with parameters $\phi$ such that $\mathit{G}_{\phi}(\mathit{X}) \rightarrow \mathit{Y}$. To aid in segmentation, an auxiliary classification task is manufactured by creating a set of class labels $\mathit{C}$ such that $\mathit{C} = \{\max(\mathit{y}_i)\}$ where $\mathit{y}_i = \{0, 1\}^{256\times256}$ for $i = 1, \ldots, N$.

\paragraph*{Auxiliary Classification:}
For the classification branch of our SAM-Mix model, we first define a classifier $R$ based on a ResNet38 \cite{wu2019wider} backbone. The primary role of this classifier is to predict class label for each input image while also providing feature maps for GradCAM generation. The classifier consists of an encoder $C_E$ and a classification head. The encoder is responsible for extracting a feature map $F_{cls} \in \mathbb{R}^{D \times h \times w}$ from the input image $X \in \mathbb{R}^{1 \times H \times W}$. 
\begin{equation}
    F_{cls} = C_E(X).
\end{equation}
After obtaining the feature maps from the last convolution layer ($F_{last}$) in the encoder, the classification head applies global average pooling ($GAP$) to produce the class logits $\hat{c} \in \mathbb{R}^{2}$.
\begin{equation}
    \hat{c} = GAP(F_{last}).
\end{equation}
These class logits are then passed to a fully connected layer for final classification. For binary classification, we minimize a binary focal loss as:
\begin{equation}
    \mathcal{L}_{cls} = -\alpha (1 - p_t)^\gamma \log(p_t),
\end{equation}
where $p_t$ is the model's predicted probability for the actual class label, $\alpha$ is a weighting factor for balancing the importance of positive/negative examples, and $\gamma$ is the focusing parameter that adjusts the rate at which easy examples are down-weighted.

To generate GradCAMs, we compute a weighted sum of the feature maps from the final convolutional layer based on the learned class weights. Specifically, for a given class, $c$, we calculate the GradCAM $g$ as:
\begin{equation}
    g = \sum_{i} w_{i}^{(c)} F_{last, i},
\end{equation}
where $F_{last, i}$ are the individual feature maps from the final convolutional block and $w_{i}^{(c)}$ are the weights from the fully connected layer corresponding to class $c$.

\paragraph*{Automated Prompt Generation:}
After generation, a GradCAM output undergoes thresholding to identify the areas with the most significant activations in the feature map. Via this process, a binary mask $b$ is generated from the GradCAM, with the activation value of each pixel in the GradCAM being compared against a predefined threshold $\tau$. If any pixel activation exceeds this threshold, it is marked as "1" in the binary mask, indicating a region of high relevance. Likewise, pixels whose activations do not exceed the threshold are marked as "0", meaning they are less relevant or otherwise inactive. 
\begin{equation}
    b(x_{coord}, y_{coord}) = 
                \begin{cases} 
                1 & \text{if } g(x, y) \geq \tau \\
                0 & \text{if } g(x, y) < \tau ,
                \end{cases}
\end{equation}
where $\tau$ is set as a percentage of the maximum activation value, such that:
\begin{equation}
    \tau = \omega \cdot \max(g),
\end{equation}
with $\omega$ being the predefined thresholding factor that determines the sensitivity. For simplicity in the binary segmentation task, we set the threshold $\tau$ to 0.5.

Once binary masks are created from the GradCAMs, the model calculates bounding boxes that define a region of interest (ROI) around each activated area. Specifically, for each activated region, the bounding box $bbox$ is calculated as

\begin{equation}
    bbox = (x_{\text{min}}, y_{\text{min}}, x_{\text{max}}, y_{\text{max}}),    
\end{equation}
where
\begin{equation}
\begin{aligned}
    x_{\text{min}} &= \min \{ x_{coord} \mid b(x_{coord}, y_{coord}) = 1 \} \\
    x_{\text{max}} &= \max \{ x_{coord} \mid b(x_{coord}, y_{coord}) = 1 \} \\
    y_{\text{min}} &= \min \{ y_{coord} \mid b(x_{coord}, y_{coord}) = 1 \} \\
    y_{\text{max}} &= \max \{ y_{coord} \mid b(x_{coord}, y_{coord}) = 1 \}.
\end{aligned}
\end{equation}

This bounding box captures the minimum and maximum coordinates along both spatial dimensions, creating a rectangular region that encloses the most relevant areas in the input image.

The bounding boxes generated during this process serve as prompts for the SAM model, automatically guiding the segmentation to focus on the ROIs identified by the GradCAM. This targeted approach enables SAM to leverage information from the classification branch, aligning the segmentation with the model's classification-based activations. This functionality allows SAM to prioritize high-saliency areas for more accurate segmentations without requiring manual or semi-automated prompt generations.

\paragraph*{Segmentation:}
The segmentation branch of SAM-Mix uses the image encoder from SAM to produce feature embeddings, which capture necessary spatial information for the segmentation task. The prompt encoder generates sparse embeddings for ROIs indicated by the automatically created bounding box prompts. The sparse embeddings focus on key points or regions within the image. On the other hand, the dense embeddings via the image encoder provide a broader view ensuring more comprehensive segmentation coverage in the image.

To ensure computational efficiency, we leverage an adapter incorporating the low-rank adaptation (LoRA) \cite{hu2021lora} into the image encoder. Specifically, it is applied to the attention mechanisms of SAM's vision transformer (ViT) base. Instead of fine-tuning the entire model, LoRA enables efficient task-specific adaptation by introducing two learnable low-rank matrices $A$ and $B$. These matrices have dimensions ($r$, $d$) and ($d$, $r$) where $d$ is the dimensionality of the input, and $r$ is the rank of the adaptation. We choose the rank $r=8$. Multiplying the matrices results in a matrix of size ($d$, $d$), maintaining the original model's capacity while learning new representations for the segmentation task. The use of LoRA significantly reduces the number of trainable parameters in the model.

After generating the sparse and dense embeddings, the SAM's decoder takes these embeddings and predicts the segmentation mask. The predicted masks are refined using the spatial information from the encoder. We employ a Dice-based loss function to optimize the model parameters.
\begin{equation}
\mathcal{L}_{\text{seg}} = 1 - \frac{2 \sum_{i} \hat{y}_i y_i}{\sum_{i} \hat{y}_i + \sum_{i} y_i},
\end{equation}
where $\mathcal{L}_{\text{seg}}$ is the segmentation loss and $\hat{y}$ is the predicted segmentation.

\section{Experimental Evaluation}
\label{sec:evaluation}

\subsection{Data}
We validate our proposed SAM-Mix method with the publicly available Liver Tumor Segmentation (LiTS) Benchmark \cite{bilic2023liver}. We split the dataset into training (100 scans), validation (5 scans), and test (25 scans) sets, with 11,437, 1,139, and 4,827 slices respectively in each set. The binary labels are then generated based on the liver's absence (0) or presence (1) in the slices. To evaluate the generalizability of SAM-Mix, we use 1,324 slices extracted from 20 CT scans in the TotalSegmentator \cite{wasserthal2023totalsegmentator} dataset.

\subsection{Implementation Details}
\textbf{Inputs:} All image slices are 0–1 normalized and reshaped to 256 × 256 × 1 before passing them to the models. We further preprocess the images by window-leveling with a width of 400 HU and a center of 40 HU. Since the liver organ spans across a relatively smaller number of slices in a scan, we extract only the middle 30\% slices to avoid class imbalance. \textbf{Baselines:} For baseline comparisons, we use fully supervised versions of U-Net~\cite{ronneberger2015u}, nnU-Net~\cite{isensee2021nnu}, TransUNet ~\cite{chen2024transunet}. We also compare SAM-Mix against another multitasking SSL method MultiMix~\cite{haque2021multimix}. For a fair comparison, the proxy labeling of MultiMix was excluded. Note that, all the baseline models are randomly initialized. We further compare our method by training and evaluating the performance of the baseline SAM as postprocessing (SAM-PP) in zero-shot and few-shot settings. To do this, we first generate the bounding box prompts and train the SAM model separately. Unlike SAM-PP, our SAM-Mix model is trained in an end-to-end manner. For both SAM-PP and SAM-Mix, we use 5, 50, and 100 labeled 2D slices to refine the bounding box prompts.\textbf{Training:} Each fully supervised model was trained for 100 epochs with a cosine-annealing learning rate scheduler and warm restarts after every 10 epochs. To demonstrate the effectiveness of our SAM-Mix model and the effectiveness of SAM-PP at generating accurate segmentations from GradCAM-based prompts in a much faster training time, the SAM-based models were trained for just 10 epochs. \textbf{Hyperparameters:} The models are trained using a learning rate of 0.001 and a batch size of 30. Each model was trained 5 times to avoid any bias and enhance reliability in model predictions. \textbf{Evaluation:} For evaluation of the segmentation performance, we use the Dice similarity (DS) and Hausdorff distance (HD). The average scores are reported across five iterations of each model.

\begin{table}[t]
    \centering
    \caption{Quantitative evaluation on the LiTS dataset demonstrates the superiority of our SAM-Mix over the fully-supervised and other semi-supervised models. The best and second best results are \textbf{bolded} and \underline{underlined}, respectively.}
    \label{tab:in}
    \resizebox{\linewidth}{!}{
    \begin{tabular}{l c c c}
        \toprule
        
        Model & Supervision & DS & HD \\ 
        \midrule
        U-Net & \multirow{3}{*}{Full} & 0.897 ± 0.010 & 13.496 ± 1.272 \\ 
        nnU-Net && 0.863 ± 0.034 & 21.407 ± 1.391 \\ 
        TransU-Net && 0.889 ± 0.005 & \underline{12.68 ± 2.109} \\
        \midrule
        MultiMix & \multirow{8}{*}{Limited} & 0.627 ± 0.007 & 21.470 ± 1.901 \\
        SAM-PP-0${}^*$ &&  0.441 ± 0.012 & 67.565 ± 1.477 \\
        SAM-PP-5 && 0.754 ± 0.006 & 36.521 ± 0.528\\
        SAM-PP-50 && 0.726 ± 0.005 & 30.993 ± 0.010\\
        SAM-PP-100  && 0.763 ± 0.003 & 26.535 ± 0.069\\
        SAM-Mix-5 && 0.919 ± 0.002 & 15.1413 ± 0.250 \\
        SAM-Mix-50 && \textbf{0.948 ± 0.002} & \textbf{9.842 ± 0.046}\\
        SAM-Mix-100 && \underline{0.941 ± 0.001} & 14.671 ± 0.052\\
        \bottomrule
        ${}^*${\footnotesize No segmentation training.}
    \end{tabular}
    }
\end{table}%

\begin{table}[t]
    \centering
    \caption{Cross-domain: Quantitative evaluation on the TotalSegmentator dataset demonstrates the superiority of our proposed SAM-Mix over the fully-supervised and other semi-supervised models. The best and second best results are \textbf{bolded} and \underline{underlined}, respectively.}
    \label{tab:cross}
    \resizebox{\linewidth}{!}{
    \begin{tabular}{lc c c c}
        \toprule
        
        Model & Supervision & \phantom{a} & DS & HD \\ 
        \midrule
        U-Net & \multirow{3}{*}{Full} && 0.669 ± 0.055 & 29.588 ± 0.052\\ 
        nnU-Net &&& 0.645 ± 0.013 & 38.497 ± 0.043\\ 
        TransU-Net &&& 0.642 ± 0.024 & 42.796 ± 0.038\\
        \midrule
        MultiMix & \multirow{8}{*}{Limited} && 0.159 ± 0.196 & 97.982 ± 0.057\\
        SAM-PP-0${}^*$ &&& 0.334 ± 0.011 & 72.150 ± 1.250  \\
        SAM-PP-5 &&& 0.593 ± 0.007 & 54.784 ± 0.545\\
        SAM-PP-50 &&& 0.579 ± 0.008 & 46.513 ± 0.035\\
        SAM-PP-100  &&& 0.605 ± 0.004 & 40.328 ± 0.068\\
        SAM-Mix-5 &&& 0.807 ± 0.002 & 17.568 ± 0.086\\
        SAM-Mix-50 &&& \textbf{0.923 ± 0.004} & \textbf{11.164 ± 0.021} \\ 
        SAM-Mix-100 &&& \underline{0.921 ± 0.001} & \underline{12.926 ± 0.023} \\
        \bottomrule
         ${}^*${\footnotesize No segmentation training.}
    \end{tabular}
    }
\end{table}%

\begin{figure}[t]
    \centering
    \includegraphics[width=\linewidth]{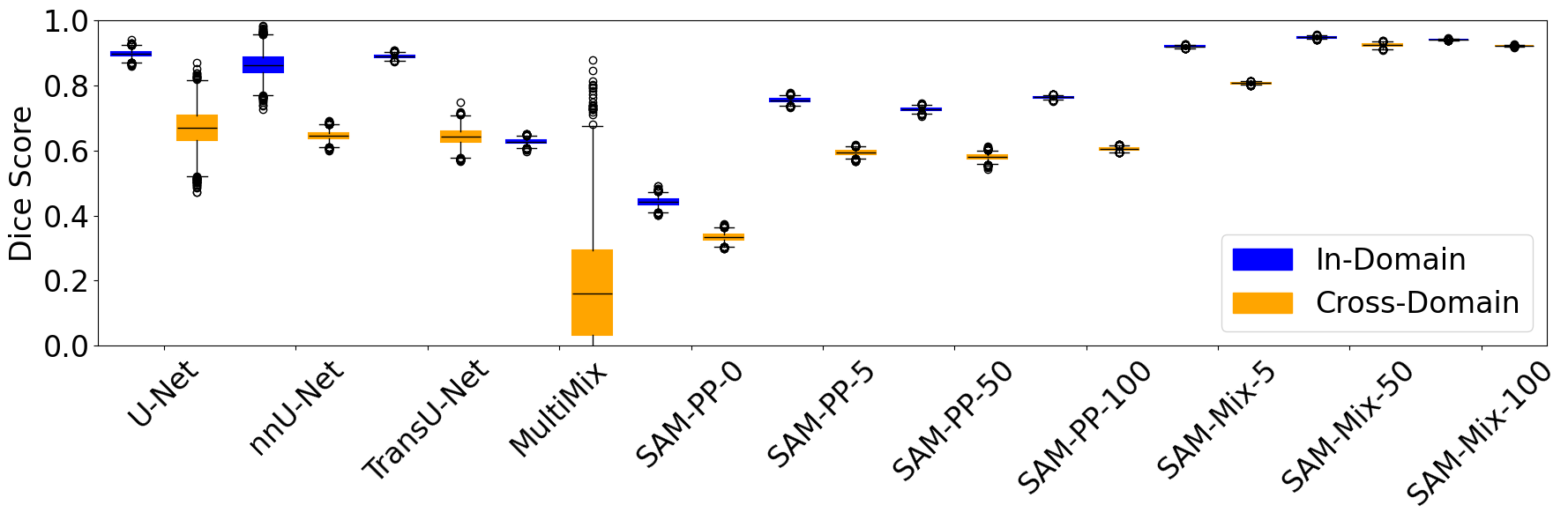}
    \caption{Dice score Box plots across the in-domain and cross-domain demonstrate the superior generalizability of SAM-Mix even at scarce labeled settings.}
    \label{fig:boxplot}
\end{figure}

\begin{figure}[t]
    \centering
    \resizebox{\linewidth}{!}
    {
    \begin{tabular}{c c c c c}
     \includegraphics[width=0.19\linewidth]{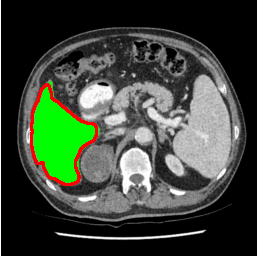}
     &
     \includegraphics[width=0.19\linewidth]{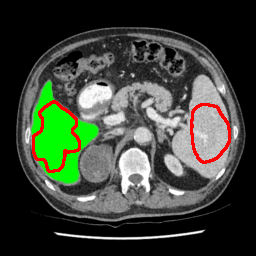}
     &
     \includegraphics[width=0.19\linewidth]{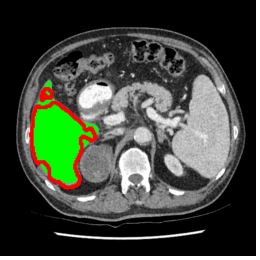}
     &
     \includegraphics[width=0.19\linewidth]{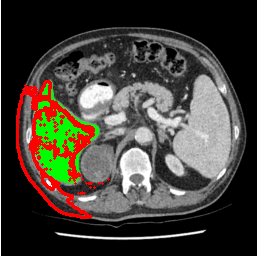}
     &
     \includegraphics[width=0.19\linewidth]{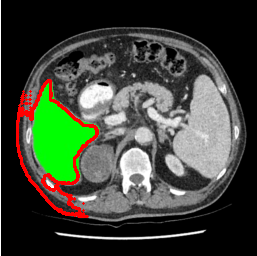}
     \\
     U-Net & nnU-Net & TransU-Net &
     SAM-PP-0 & SAM-PP-5
     \smallskip\\
     \includegraphics[width=0.19\linewidth]{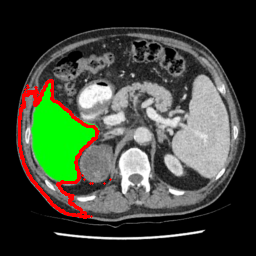}
     &
     \includegraphics[width=0.19\linewidth]{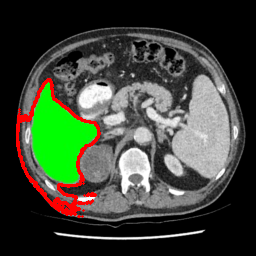}
     &
     \includegraphics[width=0.19\linewidth]{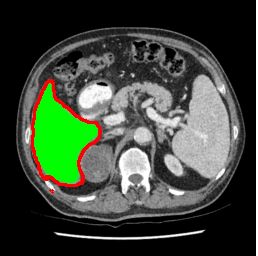}
     &
     \includegraphics[width=0.19\linewidth]{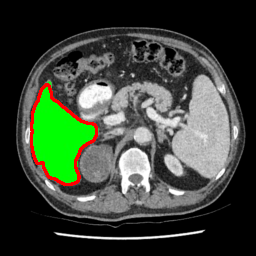}
     &
     \includegraphics[width=0.19\linewidth]{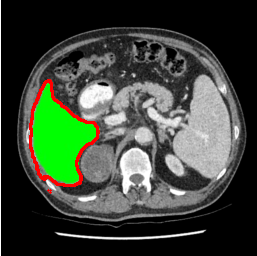}
     \\
    SAM-PP-50 & SAM-PP-100 &
     SAM-Mix-5 & SAM-Mix-50 & SAM-Mix-100
    \end{tabular}
    }
    \caption{Qualitative comparison demonstrates the superiority of SAM-Mix over other models in segmenting CT liver even when it's trained only on 5 segmentation labels. Color code: Green - Ground Truth mask, Red-predicted contour.}
    \label{fig:masks}
\end{figure}

\subsection{Results and Discussion}

\paragraph*{In-Domain Segmentation:}
Our primary findings comparing the proposed model to baseline fully-supervised models when segmenting the liver from the LiTS dataset are reported in Table~\ref{tab:in}. The reported results demonstrate that our SAM-Mix model consistently achieves higher Dice scores compared to the fully supervised baselines as well as the two-stage SAM-PP method. Against U-Net, the best-performing fully supervised method, the SAM-Mix variant trained on 50 segmentation labeled slices (SAM-Mix-50), achieves a Dice score improvement of 5.9\%. In terms of Hausdorff distance, while the fully-supervised baselines do slightly outperform SAM-Mix-5 and SAM-Mix-100, SAM-Mix-50 does achieve a lower Hausdorff distance by 22.38\%. Qualitative evaluation as shown in Figs.~\ref{fig:masks}  further affirms the superiority of SAM-Mix over baseline and existing fully-supervised methods as well as the two-stage SAM-PP variants. Furthermore, the boxplot visualization in Fig.~\ref{fig:boxplot} showcases consistently improved performance by SAM-Mix outperforming all the fully supervised and semi-supervised methods.

\begin{figure}[t]
    \centering
    \resizebox{\linewidth}{!}
    {
    \begin{tabular}{c c c c c}
     \includegraphics[width=0.19\linewidth]{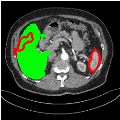}
     &
     \includegraphics[width=0.19\linewidth]{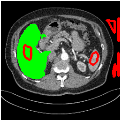}
     &
     \includegraphics[width=0.19\linewidth]{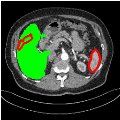}
     &
     \includegraphics[width=0.19\linewidth]{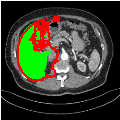}
     &
     \includegraphics[width=0.19\linewidth]{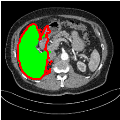}
     \\
     U-Net & nnU-Net & TransU-Net &
     SAM-PP-0 & SAM-PP-5
     \smallskip\\
     \includegraphics[width=0.19\linewidth]{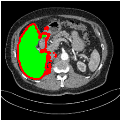}
     &
     \includegraphics[width=0.19\linewidth]{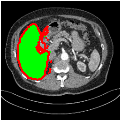}
     &
     \includegraphics[width=0.19\linewidth]{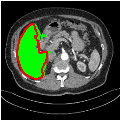}
     &
     \includegraphics[width=0.19\linewidth]{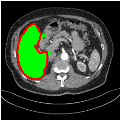}
     &
     \includegraphics[width=0.19\linewidth]{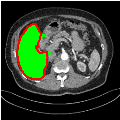}
     \\
    SAM-PP-50 & SAM-PP-100 &
     SAM-Mix-5 & SAM-Mix-50 & SAM-Mix-100
    \end{tabular}
    }
    \caption{Qualitative comparison demonstrates the superiority of SAM-Mix over other models in generalizing to cross-domain segmentation tasks.}
    \label{fig:masks2}
\end{figure}

\paragraph*{Cross-Domain Generalization:}
To test the generalizability of SAM-Mix, we validate it on a cross-domain dataset (TotalSegmentator). As reported in Table~\ref{tab:cross}, the segmentation performance of SAM-Mix on the cross-domain task is as promising as the in-domain results. Despite the data shifts, from LiTS to TotalSegmentator, SAM-Mix outperforms all of the baseline models in terms of generalizability, by significant margins. SAM-Mix-50 achieves a Dice score improvement of 25.4\%, and a lower Hausdorff distance by 62.26\% compared to the best-performing full-supervised U-Net model. These results further validate the few-shot capabilities and establish our SAM-Mix as a well-generalized tool for medical image segmentation.

\section{Conclusions}
We have presented a novel multitask learning approach to annotation-efficient medical image segmentation (SAM-Mix). 
Through an innovative auxiliary classification prediction, segmentation prompts are generated automatically. This annotation-efficient task guidance enables SAM to accurately segment liver from abdominal CT images requiring no manual or semi-automated prompting. Experimental evaluations reveal that SAM-Mix significantly improves in performance compared to various baselines for binary segmentation tasks in abdominal CT scans. Future work will explore the efficacy of incorporating newer and domain-specific versions of SAM into the SAM-Mix architecture. 

\balance
\bibliographystyle{IEEEbib}
\bibliography{refs}

\end{document}